\documentclass[final,3p,times,twocolumn]{elsarticle}

\usepackage{graphicx}
\usepackage{amsmath}
\usepackage{amsfonts}
\usepackage[colorlinks=True]{hyperref}
\usepackage[all]{hypcap}
\usepackage{comment}
\usepackage{lineno}
\modulolinenumbers[1]
\usepackage{booktabs}
\usepackage{multirow}

\journal{Journal Name}

\biboptions{numbers,sort&compress}

\begin{document}

\begin{frontmatter}

\title{ReLaText: Exploiting Visual Relationships for Arbitrary-Shaped Scene Text Detection with Graph Convolutional Networks}

\author[address1,address2]{Chixiang Ma\fnref{myfootnote}\corref{chixma}}
\ead{chixiangma@gmail.com}

\author[address2]{Lei Sun}
\ead{lsun@microsoft.com}

\author[address3,address2]{Zhuoyao Zhong\fnref{myfootnote}}
\ead{zhuoyao.zhong@gmail.com}

\author[address2]{Qiang Huo}
\ead{qianghuo@microsoft.com}

\address[address1]{Dept. of EEIS, University of Science and Technology of China, Hefei, 230026, China}
\address[address2]{Microsoft Research Asia, Beijing, 100080, China}
\address[address3]{School of EIE, South China University of Technology, Guangzhou, 510641, China}
\fntext[myfootnote]{This work was done when Chixiang Ma and Zhuoyao Zhong were interns in Speech Group, Microsoft Research Asia, Beijing, China.}
\cortext[chixma]{Corresponding author}

\begin{abstract}
We introduce a new arbitrary-shaped text detection approach named ReLaText by formulating text detection as a visual relationship detection problem. To demonstrate the effectiveness of this new formulation, we start from using a ``link'' relationship to address the challenging text-line grouping problem firstly. The key idea is to decompose text detection into two subproblems, namely detection of text primitives and prediction of link relationships between nearby text primitive pairs. Specifically, an anchor-free region proposal network based text detector is first used to detect text primitives of different scales from different feature maps of a feature pyramid network, from which a text primitive graph is constructed by linking each pair of nearby text primitives detected from a same feature map with an edge. Then, a Graph Convolutional Network (GCN) based link relationship prediction module is used to prune wrongly-linked edges in the text primitive graph to generate a number of disjoint subgraphs, each representing a detected text instance. As GCN can effectively leverage context information to improve link prediction accuracy, our GCN based text-line grouping approach can achieve better text detection accuracy than previous text-line grouping methods, especially when dealing with text instances with large inter-character or very small inter-line spacings. Consequently, the proposed ReLaText achieves state-of-the-art performance on five public text detection benchmarks, namely RCTW-17, MSRA-TD500, Total-Text, CTW1500 and DAST1500. 
\end{abstract}

\begin{keyword}
Arbitrary-Shaped Text Detection \sep Graph Convolutional Network \sep Link Prediction \sep Visual Relationship Detection
\end{keyword}

\end{frontmatter}


\section{Introduction}
\label{sec:intro}
Scene text detection has received considerable attention from computer vision and document analysis communities recently \citep{he2017deep, liao2017textboxes, liu2017deep, shi2017detecting, zhou2017east}, due to its important role in many content-based visual intelligent applications like image retrieval, autonomous driving and OCR translation. Unlike traditional OCR techniques that only deal with texts in scanned document images, scene text detection tries to detect arbitrary-shaped texts from complex natural scene images, in which texts usually occur on street nameplates, store signs, restaurant menus, product packages, advertising posters, etc. Due to high variations of text font, color, scale, orientation and language, extremely complex backgrounds, as well as various distortions and artifacts caused by image capturing like non-uniform illumination, low contrast, low resolution, and occlusion, scene text detection is still an unsolved problem.

\begin{figure}[t]
    \centering
    \includegraphics[width=\linewidth]{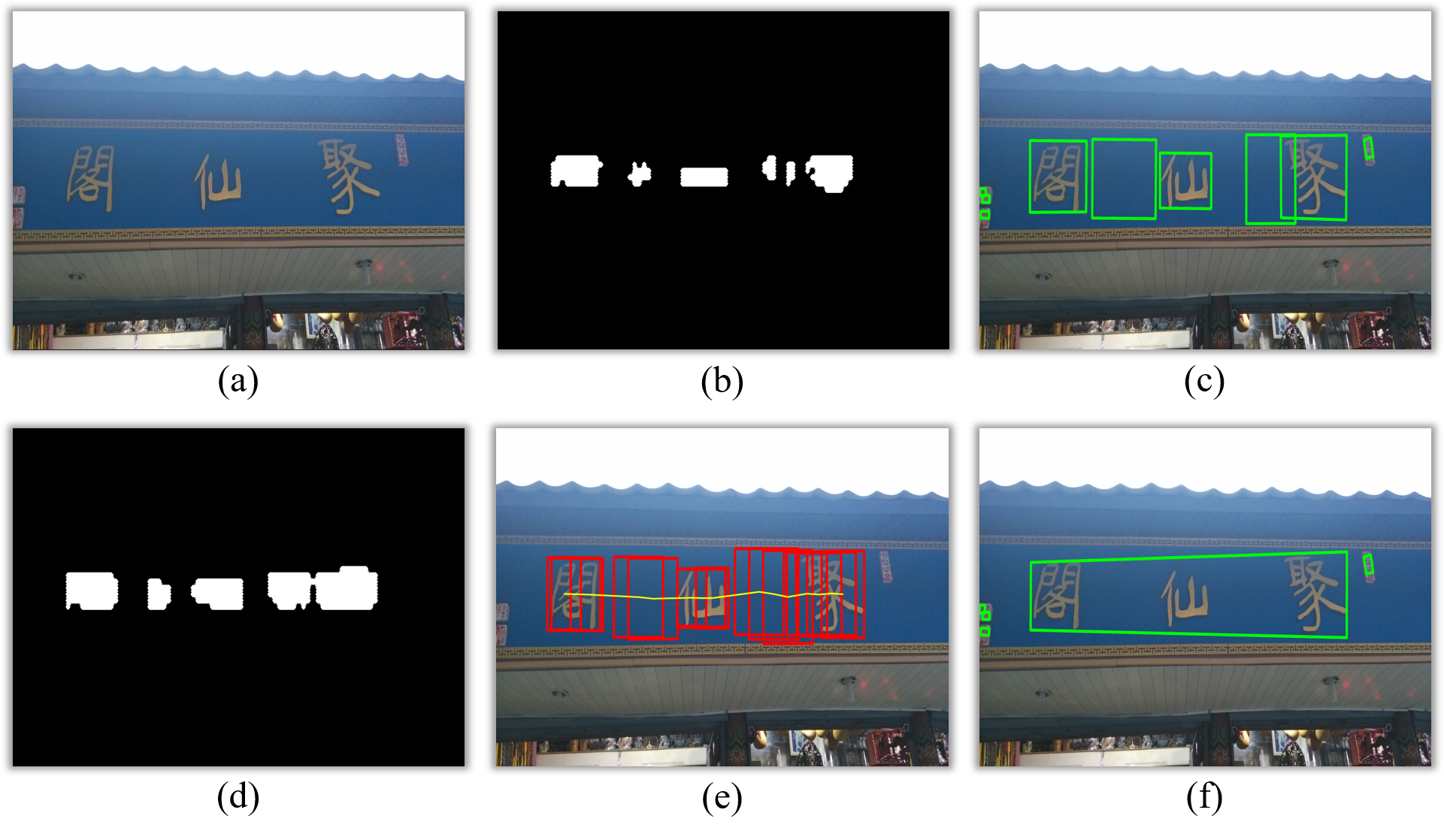}
    \caption{(a) Original image from RCTW-17; (b) Binarized textness score map from SegLink; (c) Detection results of SegLink; (d) Binarized textness score map from our method; (e) Grouped text primitives of our method based on predicted link relationships; (f) Detection results of our method.}
    \label{fig:motivation}
\end{figure}

In recent years, convolutional neural network (CNN) based scene text detection approaches have made great progress and substantially outperformed traditional MSER or SWT based text detection methods (e.g., \citep{neumann2010method, yin2013robust, epshtein2010detecting}) in terms of both accuracy and capability. As earlier scene text benchmark datasets like ICDAR'13, '15, '17\footnotemark\footnotetext{\url{https://rrc.cvc.uab.es/}}~do not contain curved texts, most earlier methods \citep{he2017deep, liao2017textboxes, liu2017deep, shi2017detecting, zhou2017east, sun2015robust, zhang2016multi, jaderberg2016reading, gupta2016synthetic, ma2018arbitrary, tian2016detecting, lyu2018multi, liu2018fots, zhong2019improved, he2020realtime} simply assume that texts are arranged in a straight line manner and adapt CNN-based object detection and segmentation frameworks, like Faster R-CNN \citep{ren2015faster}, SSD \citep{liu2016ssd}, YOLO \citep{redmon2016you}, DenseBox \citep{huang2015densebox} and FCN \citep{long2015fully}, to solve the text detection problem. Although these methods can deal with horizontal and multi-oriented text-lines well, they are incapable of detecting curved texts. Recently, with the emergence of two curved-text focused datasets, namely Total-Text \citep{ch2017total} and CTW1500 \citep{liu2019curved}, the research focus of text detection has shifted from horizontal or multi-oriented scene texts to more challenging curved or arbitrary-shaped scene texts. This has spawned lots of more effective approaches which can be roughly classified into two categories: top-down approaches and bottom-up approaches. Top-down approaches usually use Region Proposal Network (RPN) \citep{ren2015faster} to generate coarse rectangular proposals for possible curved text instances firstly, then either modify the bounding box regression module of the second stage detector to predict a tighter polygon-shaped bounding box \citep{liu2019curved, wang2019arbitrary} or add a mask prediction module to predict a segmentation mask \citep{lyu2018mask, xie2019scene, huang2019mask, zhang2019look} for the corresponding text instance in each positive proposal. Although these approaches, especially the Mask R-CNN \citep{he2017mask} based ones \citep{xie2019scene, huang2019mask}, have achieved superior performance on some benchmark datasets, they are not robust to nearby long curved text instances which appear often in some commodity images, e.g., the DAST1500 dataset \citep{tang2019detecting}. This is because the rectangular proposals of nearby long curved text instances are highly overlapped, which will cause some of them to be wrongly suppressed by the non-maximum suppression (NMS) algorithm so that the corresponding text instances cannot be detected correctly. Unlike top-down approaches, bottom-up approaches can get rid of the limitation of rectangular proposals. These approaches either detect candidate text segments (characters or parts of words/text-lines) or predict a pixel-level textness score map firstly, then use different methods to group detected text segments or pixels into words/text-lines and calculate the corresponding bounding boxes. The difficulties lie in how to group the detected text segments or pixels into words/text-lines robustly. Segment based methods have tried rule and character embedding based line grouping algorithms to solve this problem (e.g., \citep{tian2016detecting, hu2017wordsup, liu2018detecting}), but their results on curved text datasets are still worse than pixel-based methods. Pixel-based methods directly use local pixel connectivity to merge pixels on binarized textness score maps into words/text-lines. Although the pixel merging accuracy has been improved significantly by leveraging various auxiliary information (e.g., \citep{shi2017detecting, wu2017self, xu2019textfield, xue2019msr}), these methods cannot detect text instances with large inter-character spacings robustly because pixels within large inter-character spacings tend to be misclassified as non-text (Fig.~\ref{fig:motivation}(b)) so that the text-line is over-segmented (Fig.~\ref{fig:motivation}(c)).

To address the above problems, we propose to solve the text detection problem from a new perspective by formulating text detection as a visual relationship detection problem instead of an object detection or segmentation problem. In the field of visual relationship detection, visual relationships are defined as $\langle{subject, predicate, object}\rangle$ triplets, where the ``subject'' is related to the ``object'' by the ``predicate'' relationship. A typical paradigm for this task is composed of three modules, i.e., individual object detection, subject-object pair construction and relationship classification. Text detection can be easily formulated as a visual relationship detection problem, namely, the ``subject'' or ``object'' can be defined as a text primitive which represents a text segment or a whole word/text-line, and the ``predicate'' can be defined as some relationships between text primitive pairs. We argue that visual relationship detection is a more flexible problem formulation for text detection. Because, with this new formulation, some existing difficult problems in text detection like text-line grouping and duplicate removal can be solved in a unified framework by formulating them as some specific relationship prediction problems between text primitive pairs. For example, for text-line grouping, a ``link'' relationship can be defined to indicate whether two text segments belong to a same word/text-line. To solve the duplicate removal problem, a ``duplicate'' relationship can be defined to determine whether two overlapped bounding boxes correspond to a same word/text-line. Moreover, the visual relationship detection framework can also be applied in other document understanding tasks like key-value pair extraction from documents \citep{davis2019deep} and table recognition \citep{qasim2019rethinking}. With a same formulation, these systems can be seamlessly integrated together making it much easier to leverage multi-task learning to improve the accuracy for each task.

In this work, to demonstrate the effectiveness of this new formulation, we start from using a ``link'' relationship to address the challenging text-line grouping problem firstly. To this end, we decompose text detection into two key subproblems, namely detection of text primitives and prediction of link relationships between text primitive pairs. Based on the predicted link relationships, text primitives are grouped into text instances. In our conference paper \citep{ma2019relation}, we have demonstrated that leveraging link relationships predicted by a relation network \citep{zhang2017relationship} to do text-line grouping can achieve better text detection accuracy than previous local pixel connectivity based text-line grouping methods. In this paper, we introduce further a new Graph Convolutional Network (GCN) based link prediction approach \citep{kipf2017semi}, which can leverage context information more effectively than relation network to improve the link prediction accuracy so that even text instances with large inter-character spacings or very small inter-line spacings can be robustly detected by our text detector. Consequently, the proposed text detector, ReLaText, achieves state-of-the-art performance on five public text detection benchmarks, namely RCTW-17, MSRA-TD500, Total-Text, CTW1500 and DAST1500.

Although the preliminary results of leveraging link relationship prediction to solve the text-line grouping problem have been reported in our conference paper \citep{ma2019relation}, we extend it in this paper significantly in the following aspects: (1) We adopt GCN to replace relation network to do link relationship prediction and achieve improved text detection accuracy; (2) More ablation studies are conducted to demonstrate the effectiveness of our GCN based link relationship prediction approach; (3) Related works on scene text detection, visual relationship detection and graph convolutional network are reviewed more comprehensively; (4) Experimental results on two more public benchmark datasets, namely RCTW-17 and DAST1500, are presented to compare our approach with other state-of-the-art approaches.

\section{Related Work}
\label{sec:related_work}
\subsection{Scene Text Detection}
\label{subsec:STD}
Before the deep learning era, there were only a few works paying attention to arbitrary-shaped text detection. Shivakumara et al. \cite{shivakumara2013curve} proposed a quad tree based approach to detecting curved texts from videos.  Fabrizio et al. \cite{fabrizio2016textcatcher} proposed to group extracted candidate text CCs into a graph, in which arbitrary-shaped text lines are detected based on some regularity properties. As the performance of these methods heavily depends on heuristic rules or handcrafted features, they are not as robust as recent deep learning based approaches.

With the rapid development of deep learning, numerous CNN based text detection methods have been proposed and substantially outperformed traditional methods by a big margin in terms of both accuracy and capability. These methods can be roughly classified into two categories: top-down methods and bottom-up methods.
\\
\textbf{Top-down methods}. Top-down methods typically treat text as a special kind of object, and directly adapt state-of-the-art top-down object detection or instance segmentation frameworks to solve the text detection problem. Jaderberg et al. \cite{jaderberg2016reading} adopted R-CNN \citep{girshick2014rich} for text detection first, but its performance was limited by the traditional region proposal generation methods \citep{gomez2017textproposals}. Later, Zhong et al. \cite{zhong2017deeptext}, Liao et al.\cite{liao2017textboxes} and Gupta et al. \cite{gupta2016synthetic} adopted Faster R-CNN, SSD and YOLO to detect horizontal texts respectively. To extend Faster R-CNN and SSD to multi-oriented text detection, Ma et al. \cite{ma2018arbitrary} and Liu et al. \cite{liu2017deep} proposed to utilize rotated rectangular or quadrilateral anchors to hunt for inclined text proposals. Since directly predicting the vertex coordinates of quadrilateral bounding boxes suffers from a label confusion issue about the vertex order, Liu et al. \cite{liu2019omnidirectional} proposed to discretize the bounding box into key edges and learn the correct match-type with a multi-class classifier. Moreover, as the anchor mechanism used by Faster R-CNN and SSD is inflexible for text detection tasks, Zhou et al. \cite{zhou2017east} and He et al. \cite{he2017deep} followed the ``anchor-free'' idea of DenseBox \citep{huang2015densebox} and proposed to use an FCN \citep{long2015fully} to directly output the pixel-wise textness scores and bounding boxes of the concerned text instances through all locations and scales of an image. Although more flexible, the capabilities of DenseBox-based one-stage text detectors are limited because they cannot detect long or large text instances effectively \citep{zhou2017east}. To address this issue, Zhong et al. \cite{zhong2019anchor} proposed to use DenseBox to replace the original anchor-based RPN in Faster R-CNN so that their Faster R-CNN based text detector can get rid of the limitations of anchor mechanism while preserving good accuracy for multi-oriented text detection. Another method \citep{lyu2018multi} first generated candidate boxes by sampling and grouping the detected corner points of text bounding boxes, among which unreasonable boxes were eliminated by position sensitive segmentation scores.  Since the rectangular or quadrilateral bounding boxes predicted by the above-mentioned text detectors cannot enclose curved texts tightly enough, these methods cannot detect curved texts effectively. To extend R-FCN \citep{dai2016rfcn} to curved text detection, Liu et al. \cite{liu2019curved} modified the bounding box regression module to predict a tighter polygonal bounding box with 14 points for each text proposal, which is further refined by a recurrent neural network to make the boundary more accurate. Wang et al. \cite{wang2019arbitrary} argued that polygons of fixed 14 points were not precise enough for long curved text lines, so they proposed to use a recurrent neural network to predict polygons of different numbers of points for texts of different shapes. Meanwhile, another category of methods \citep{lyu2018mask, xie2019scene, huang2019mask, zhang2019look} formulated text detection as an instance segmentation problem and borrowed existing top-down instance segmentation frameworks like Mask R-CNN \citep{he2017mask} to predict a segmentation mask and optionally extra geometric attributes for the corresponding text instance in each positive proposal. Although these methods, especially the Mask R-CNN based ones \citep{xie2019scene, huang2019mask}, have achieved superior performance on most benchmark datasets like Total-Text and CTW1500, they are not robust to nearby long curved text instances. Tang et al. \cite{tang2019detecting} introduced a new dense and arbitrary-shaped text detection dataset, i.e., DAST1500, which is mainly composed of commodity images, to demonstrate this. The main reason is that the rectangular proposals of nearby long curved text instances generated by existing top-down methods are highly overlapped, and some of them may be wrongly suppressed by the non-maximum suppression (NMS) algorithm so that the corresponding text instances cannot be detected correctly.
\\
\textbf{Bottom-up methods}. Bottom-up methods generally follow a component-grouping paradigm, i.e., detect text components first and then group these components into text instances. Compared with top-down methods, bottom-up methods can get rid of the limitations of the region proposal generation module. Based on the granularity of text components, these methods can be further divided into two categories: pixel-level methods and segment-level methods.
\\
\textbf{1) Pixel-level:} Pixel-based methods usually leverage semantic segmentation or instance segmentation frameworks to predict a pixel-level textness score map firstly, then use different methods to group text pixels into  words/text-lines  and  calculate  the  corresponding bounding  boxes. Zhang et al. \cite{zhang2016multi} first used FCN to predict text blocks from which character candidates are extracted with MSER, then post-processing methods were used to generate text-lines. More recent works in this category directly used local pixel connectivity (e.g., 8-neighbourhood) to merge pixels on binarized textness score maps into CCs, each of which represents a word/text-line. In order to avoid merging nearby words/text-lines together or over-segmenting words/text-lines into pieces, these approaches tried to leverage other auxiliary information, e.g., link prediction \citep{shi2017detecting, deng2018pixellink}, progressive scale expansion \citep{Wang2019psenet, liu2019towards}, text border prediction \citep{wu2017self}, text center line extraction \citep{long2018textsnake, xue2019msr}, text center-border probability prediction \citep{zhu2018textmountain}, Markov clustering \citep{liu2018learning}, direction field prediction \citep{xu2019textfield}, pixel embedding mapping \citep{tian2019learning} and character affinity estimation \citep{baek2019character} to enhance pixel merging performance. Although these local pixel connectivity based line grouping approaches have achieved superior performance on benchmark datasets, we find that they tend to over segment text instances with large inter-character spacings into pieces, which is also mentioned in \citep{shi2017detecting, xue2019msr, xu2019textfield}.
\\
\textbf{2) Segment-level:} Segment-based methods detect text segments firstly, each of which contains a character or part of a word/text-line. The difficulties of these approaches also lie in how to robustly group the detected text segments into words/text-lines. Earlier works in this category like CTPN \citep{tian2016detecting} and Wordsup \citep{hu2017wordsup} adopted rule-based methods to group the detected text segments into horizontal or multi-oriented text instances, which are not robust to curved texts. Recently, Liu et al. \cite{liu2018detecting} proposed a character embedding based approach to group detected characters into curved text-lines. However, their reported results on Total-Text are worse than pixel-level methods. Our proposed ReLaText is also a segment-level bottom-up approach, but we formulate text detection as a visual relationship detection problem and take advantage of the graph convolutional network to predict link relationships between text segments so that more robust text-line grouping can be achieved for arbitrary-shaped texts.

\label{sec:methodology}
\begin{figure*}[t]
    \centering
    \includegraphics[width=1.0\linewidth]{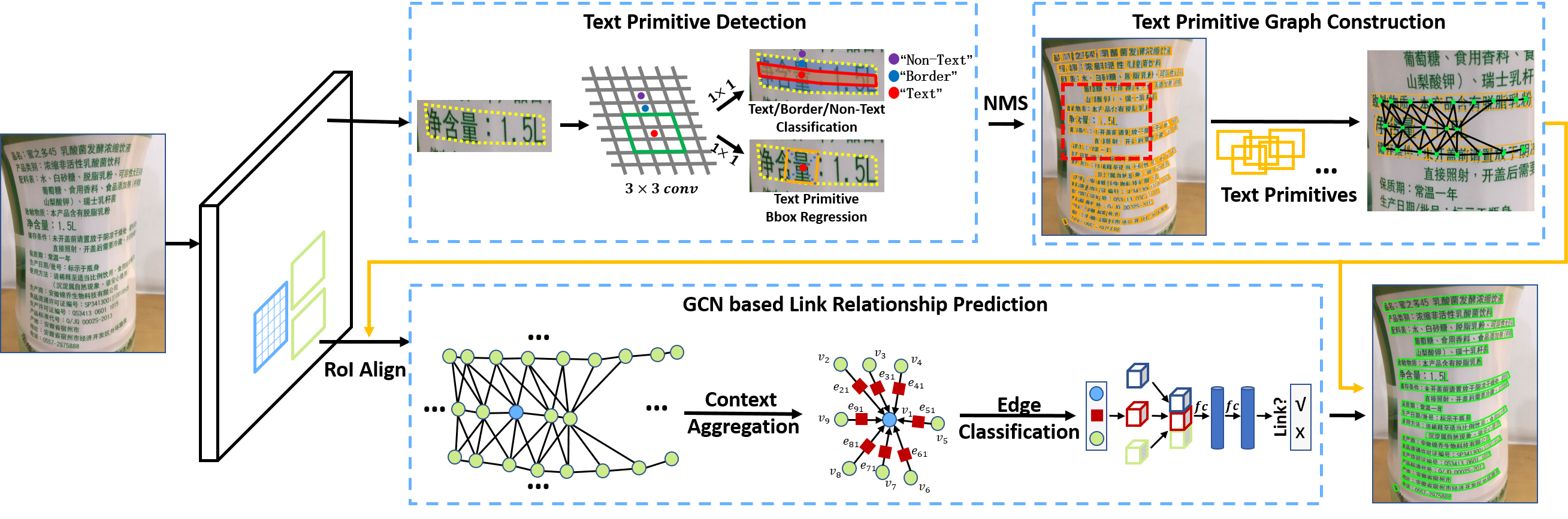}
    \caption{Overview of our ReLaText. For brevity, the FPN backbone network is omitted and only one of the three parallel forward paths on the feature pyramid levels is illustrated.}
    \label{fig:overview}
\end{figure*}

\subsection{Visual Relationship Detection}
\label{subsec:VRD}
Visual relationship detection has experienced a rapid development since some large scale datasets like VRD \citep{lu2016visual} were released. Visual relationships are defined as $\langle{subject, predicate, object}\rangle$ triplets, where the ``subject'' is related to the ``object'' by the ``predicate'' relationship. The goal of visual relationship detection is to detect objects along with predicting the relationships between object pairs from images. A typical paradigm for this task is composed of three modules, i.e., individual object detection, subject-object pair construction and relationship classification \citep{lu2016visual, li2017vip, zhu2018deep, yu2017visual, xu2017scene, yang2018graph, dai2017detecting}. As context information is important to improve relationship classiﬁcation accuracy, most previous methods tried to leverage wider context by simultaneously taking both “subject” and “object” proposals as well as their union as input to predict the “predicate” relationship \citep{lu2016visual, li2017vip, zhu2018deep, yu2017visual}. Later works \citep{xu2017scene, yang2018graph, woo2018linknet} proposed to use GCNs or its variants to further enhance the context information. Other than context information, the semantic relationships between objects and their corresponding predicates are also very important to improve accuracy \citep{lu2016visual, yu2017visual, zhuang2017towards}. To leverage this information, Lu et al. \cite{lu2016visual} leveraged language priors from semantic word embeddings to finetune the likelihood of a predicted relationship. Yu et al. \cite{yu2017visual} proposed to distill the internal and external linguistic knowledge into a deep neural network to regularize visual model learning. In this work, we propose a new GCN based link relationship prediction approach for improving link prediction accuracy.

\subsection{Graph Convolutional Network}
\label{subsec:GCN_related}
Although deep learning has revolutionized many machine learning tasks with data typically represented in the Euclidean space, there are a large number of real-world applications requiring to deal with non-Euclidean data, which is usually represented with arbitrarily structured graphs and imposes significant challenges on existing machine learning algorithms. Recently, many studies on extending deep learning approaches for graph data have emerged, and we refer readers to \citep{goyal2018graph} for a comprehensive survey. Early works first attempted to use recursive neural network to process such graph-structured data \citep{sperduti1997supervised}. Follow-up works raised an increasing interest in generalizing convolutions to the graph domain to decompose the complicated computational operations over graph-structured data into a series of local operations for each node at each time step. The advances in this direction can be divided into two categories: spectral-based approaches and spatial-based approaches. In general, spectral-based approaches \citep{bruna2014spectral, henaff2015deep} process graph data from the perspective of graph signal processing, and define the convolutional operation based on the spectral graph theory. On the contrary, spatial-based approaches \citep{atwood2016diffusion, niepert2016learning, hamilton2017inductive} define the convolutional operation directly on the graph, only involving the spatially close neighbors. This series of works are the most related to this paper. One of the challenges of these approaches is to design the graph convolutional operator to work on nodes with different degrees. Duvenaud et al. \cite{duvenaud2015convolutional} achieved it by learning an individual weight matrix for each node degree. Hamilton et al. \cite{hamilton2017inductive} adopted sampling strategy to obtain a fixed number of neighbors for each node. Later, Kipf et al. \cite{kipf2017semi} simplified the spectral-based graph convolutions with a localized first-order approximation, which bridged the gap between spectral-based approaches and spatial-based approaches. More recently, the latest advances on GCNs include attentional weighted edges \citep{velivckovic2018graph}, dynamic edge convolution \citep{wang2019dynamic}, etc.

\section{Methodology}
\subsection{Overview}
\label{subsec:overview}
We propose to represent each text instance as a sparse directed subgraph, where nodes and edges represent text primitives and link relationships between nearby text primitive pairs, respectively. Here, each text primitive represents a text segment as in CTPN. Based on this representation, our ReLaText is designed to comprise three key modules: 1) An anchor-free RPN based text detection module to detect text primitives from a feature pyramid generated by a Feature Pyramid Network (FPN) backbone \citep{lin2017feature}; 2) A text primitive graph construction module to group text primitives detected from different feature maps into different subgraphs respectively by linking each pair of nearby text primitives from a same feature map with an edge; 3) A GCN-based link relationship prediction module to prune wrongly-linked edges in all subgraphs to generate the final results. A brief pipeline of ReLaText is illustrated in Fig.~\ref{fig:overview}, and the details of each module are described in the following subsections.

\subsection{Text Primitive Detection}
\label{subsec:TPD}
\begin{figure}[t]
    \centering
    \includegraphics[width=\linewidth]{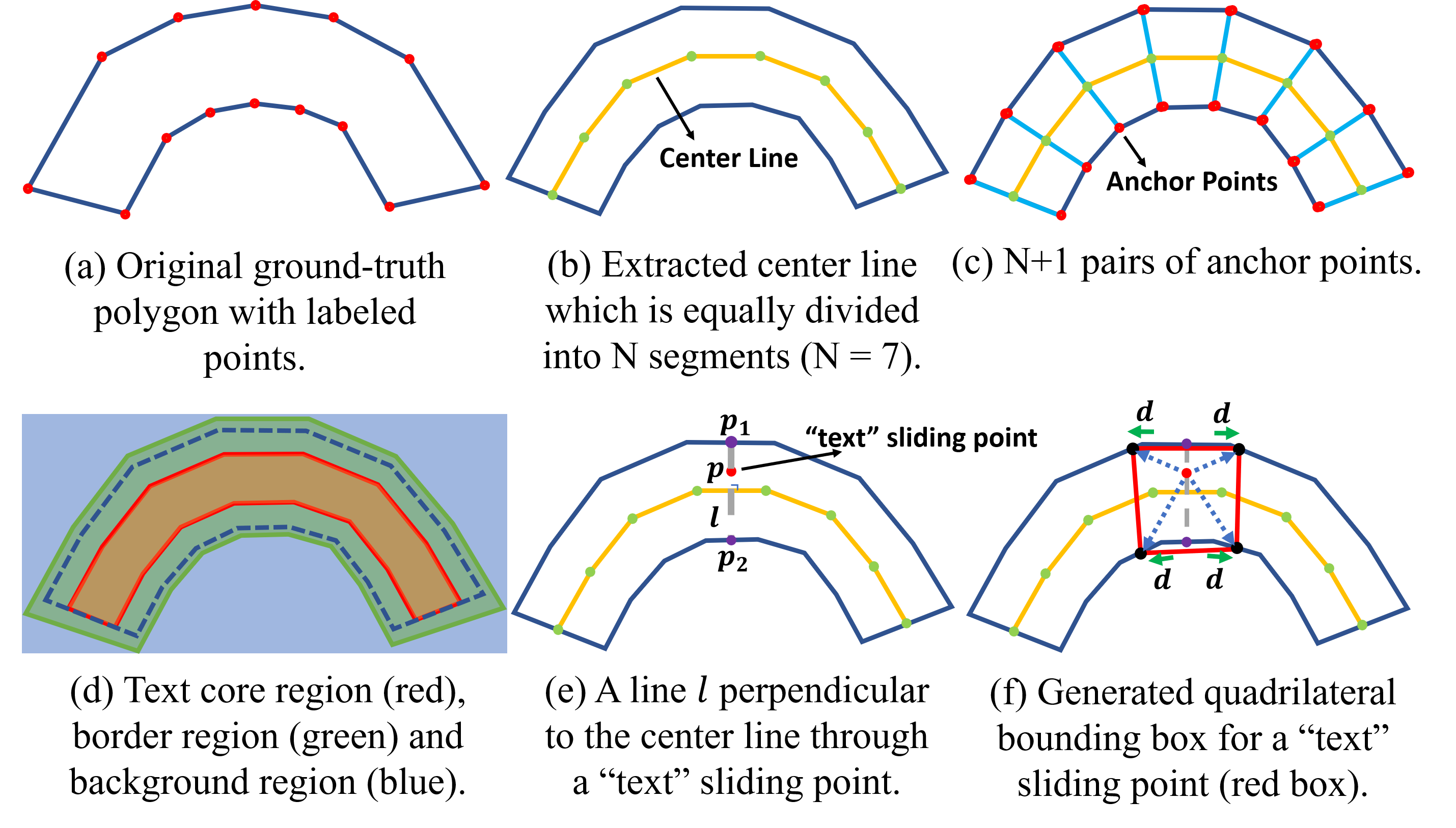}
    \caption{Illustration of ground-truth bounding box generation for a text primitive. Given a ground-truth polygon, we represent it with $N+1$ pairs of anchor points following Long et al. \cite{long2018textsnake} as shown in (a)-(c). Then, we shrink and expand the polygon by scaling factors of 0.5 and 1.2 respectively to generate a text core region and a border region (d). Only pixels within the text core region are labeled as ``text''. Let $p$ denote a ``text'' sliding point (red point in (e)) and $l$ denote a line which is perpendicular to the center line through $p$. Assume that $l$ intersects with two long curved edges of the polygon at points $p_1$ and $p_2$ (purple points in (e)). Then, we move $p_1$ and $p_2$ forward and backward along the upper and lower curved edges $d$ pixels respectively to get four vertices (black points in (f)) of the corresponding text primitive, where $d$ is set as $12$, $24$ and $81$ for $P_2$, $P_3$ and $P_4$, respectively.}
    \label{fig:gt}
\end{figure}

We adopt an Anchor-Free RPN (AF-RPN) method \citep{zhong2019anchor} to detect text primitives from a feature pyramid generated by FPN, which is built on top of ResNet-50 \citep{he2016deep}. Here, the feature pyramid consists of four levels, i.e., $P_2$, $P_3$, $P_4$ and $P_5$, whose strides are 4, 8, 16 and 32 pixels, respectively. All feature pyramid levels have $C=256$ channels. The key idea of AF-RPN is to use different DenseBox-based detection modules \citep{huang2015densebox} to detect text instances of different scales from different feature pyramid levels so that it is more robust to large text-scale variance. To define the scales of arbitrary-shaped text instances, we represent the boundary of each text instance by a polygon with a fixed number of anchor point pairs on its two long sides (Fig.~\ref{fig:gt}(a-c)). Based on this, we define the scale of a text instance as the length of its shorter side, which is estimated by the average length of lines connecting anchor point pairs (light blue lines in Fig.~\ref{fig:gt}(c)). In the training stage, each ground-truth polygon is only assigned to one feature pyramid level according to its scale and ignored by other feature pyramid levels. Then, we use three scale-specific detection modules to detect text primitives contained in small ($4$px-$23$px), medium ($24$px-$48$px) and large (\textgreater$48$px) text instances from $P_2$, $P_3$ and $P_4$, respectively.

Furthermore, we borrow the idea of border learning \citep{wu2017self} to enhance the robustness of each detection module to nearby text-lines. Specifically, for each ground-truth polygon in a raw image, we shrink its short side by a scaling factor of 0.5 to create a corresponding text core region (red region in Fig.~\ref{fig:gt}(d)). Meantime, we expand its short side by a scaling factor of 1.2 and define the region outside the text core region but inside the expanded polygon as a border region (green region in Fig.~\ref{fig:gt}(d)). Regions outside all expanded polygons are background regions. During training, each pixel on each pyramid level can be mapped back to a sliding point \citep{zhong2019anchor} in the raw image. For a pixel on a pyramid level, if its corresponding sliding point locates in the text core or border region of a ground truth polygon assigned to this pyramid level, it is labeled as ``text'' or “border", respectively. Pixels whose corresponding sliding points locate in ground truth polygons assigned to other pyramid levels are ignored. The remaining pixels are labeled as “non-text". For each “text” pixel, we use the algorithm depicted in Fig.~\ref{fig:gt}(e-f) to generate the ground-truth bounding box of its corresponding text primitive. In the inference stage, if a pixel on a pyramid level is classified as “text", its corresponding detection module will give it a “text” label and directly predict the offsets from it to the vertices of its corresponding quadrilateral text primitive. As depicted in Fig.~\ref{fig:overview}, each detection module is implemented as a $3\times3$ convolutional layer followed by two sibling $1\times1$ convolutional layers for text/border/non-text classification and quadrilateral bounding box regression, respectively.

To reduce false alarms, we only keep “text" pixels whose textness scores are higher than a pre-defined threshold which is set as 0.85 in our current implementation. After that, on each pyramid level, we use the standard NMS algorithm with an Intersection-over-Union (IoU) threshold of 0.6 to remove redundant text primitives.

\subsection{Text Primitive Graph Construction}
\label{subsubsec:SGC}
We group the detected text primitives from all pyramid levels into a directed graph which is denoted as $\textbf{G}=\{\textbf{V}, \textbf{E}\}$. $\textbf{V}=\{v_1, v_2, ..., v_N\}$ is the node set with element $v_i$ denoting the $i$-th text primitive. $\textbf{E}=\{e_{i\rightarrow{j}}=(v_i, v_j)|v_i, v_j \in \textbf{V}\}$ is the edge set with element $e_{i\rightarrow{j}}$ denoting an edge pointing from $v_i$ to $v_j$. Given $N$ nodes, there are $N(N-1)$ possible edges between all node pairs. However, if we take all these edges into account, the computation costs for the succeeding link relationship prediction step will be very expensive. Actually, it is unnecessary to predict link relationships between all text primitive pairs as each text instance can be perfectly represented by a sequence of ordered text primitives with link relationships between only nearby text primitive pairs. Moreover, as sizes of text primitives in a same text instance should be similar, we can ignore the link relationships of two text primitives detected from different pyramid levels. According to these assumptions, only pairs of nearby text primitives detected from a same pyramid level are linked with edges. Specifically, given two text primitives $v_i$ and $v_j$, we define a score to indicate whether there exists a directed edge pointing from $v_i$ to $v_j$ as follows:
\begin{equation}
    \textrm{exist}(e_{i\rightarrow{j}})=\left\{
        \begin{aligned}
        1 & , & \textrm{if}~v_i \in \textrm{KNN}(v_j), \\
        0 & , & \textrm{otherwise},
        \end{aligned}
    \right.
\end{equation}
where KNN($v_j$) denotes the set of K (K=10) nearest neighbors of $v_j$ at the same pyramid level. The Euclidean distance between the center points of two text primitives is taken as the distance measure. 

In this way, the generated text primitive graph can be divided into three disjoint subgraphs, i.e., $\textbf{G}_{\textbf{P}_\textbf{2}}$, $\textbf{G}_{\textbf{P}_\textbf{3}}$ and $\textbf{G}_{\textbf{P}_\textbf{4}}$, which are generated from three feature pyramid levels, $P_2$, $P_3$ and $P_4$, respectively. In each subgraph, there are K edges pointing to each node.

\subsection{GCN based Link Relationship Prediction}
\label{subsec:GCN}

\subsubsection{Context-enhanced Node Representation} 
\label{subsubsec:GCN}

For each detected text primitive, we extract a visual feature from its bounding box on its corresponding feature map as its initial node representation. Specifically, for each text primitive $v_i$, we adopt the RoI Align algorithm \citep{he2017mask} to extract a $256 \times 5 \times 5$ feature descriptor from its bounding box on its corresponding pyramid level firstly, which is then fed into a 2-hidden-layer fully connected neural network with 512 nodes at each layer to generate its initial node representation $\textbf{g}_i^{(0)}$.

After that, we use a graph convolutional neural network to enhance node representations by propagating context between nodes in each text primitive subgraph. Here, we make some modifications to the GCN topology proposed in \citep{kipf2017semi} to improve its capability. Specifically, for a target node $v_i$ in a subgraph $\textbf{G}_{\textbf{P}_\textbf{k}} (\textrm{k} = 2, 3, 4)$, the representations of $v_i$ and its neighboring nodes $\mathcal{N}(v_i)=\{v_j|\textrm{exist}(e_{j\rightarrow{i}})=1\}$ are first transformed via
two learned non-linear transformation functions $f_v$ and $f_e$, respectively. Then, these transformed representations are gathered with predetermined weights $\boldsymbol{\alpha}$, followed by a non-linear activation function $\sigma$. This layer-wise propagation can be written as

\begin{equation}
    \label{eq:GCN}
    \mathbf{g}_i^{(l+1)} = \sigma\left(f_v^{(l)}\left(\mathbf{g}_{i}^{(l)}\right) + \sum_{v_j \in \mathcal{N}(v_i)}\alpha_{ji}f_e^{(l)}\left(\mathbf{g}_j^{(l)}\right)\right) \;,
\end{equation}
where $\mathbf{g}_{i}^{(l)}$ is the updated node representation of $v_i$ output from the $l$-th GCN layer. $f_v^{(l)}$ and $f_e^{(l)}$ are implemented with multi-layer perceptrons (MLPs) with rectified linear unit (ReLU) between layers. To normalize the transformed neighboring representations, we set $\alpha_{ji}$ as $1/|\mathcal{N}(v_i)|$. And $\sigma(\cdot)$ is impletemented as a ReLU activation function.

According to Eq.~(\ref{eq:GCN}), each node in each graph convolutional layer can only interact with its first-order neighbors. In order to  increase the receptive field of each node, we stack $L$ graph convolutional layers to construct an $L$-layer graph convolutional network. In this way, each node can leverage the information from more distant neighbors in its corresponding subgraph to improve its node representation. The outputs of the last GCN layer are used as the final node representations, i.e., $\mathbf{g}_{i}^{(L)}$ is the final node representation of $v_i$.

\subsubsection{Edge Classification}
\label{subsec:RN}

The edge classification module is used to prune edges that link two text primitives not belonging to a same text instance. Given an edge $e_{i\rightarrow{j}}$ directed from $v_i$ to $v_j$, we extract its feature representation $\textbf{x}_{ij}$ by concatenating the node representations of $v_i$ and $v_j$ and the spatial compatibility feature of their bounding boxes $b_i$ and $b_j$, i.e., $\textbf{x}_{ij}=[\textbf{g}_i^{(L)}; \textbf{l}_{ij}; \textbf{g}_j^{(L)}]$. The spatial compatibility feature $\textbf{l}_{ij}$ is used to measure the relative scale and location relationships between the bounding boxes of $v_i$ and $v_j$. Following Zhang et al. \cite{zhang2017relationship}, let $b_{ij}$ denote the union bounding box of $b_i$ and $b_j$, then $\textbf{l}_{ij}$ is defined as a 18-d vector concatenating three 6-d vectors, which indicate the box delta of $b_i$ and $b_j$, $b_i$ and $b_{ij}$, $b_j$ and $b_{ij}$, respectively. Given two bounding boxes $b_i=\{x^i;y^i;w^i;h^i\}$ and $b_j=\{x^j;y^j;w^j;h^j\}$, their box delta is defined as $\bigtriangleup(b_i,b_j)=(t_x^{ij}, t_y^{ij}, t_w^{ij}, t_h^{ij}, t_x^{ji}, t_y^{ji})$ where each dimension is given by

\begin{align}
    \nonumber
    &t_{x}^{ij}=\left(x^{i}-x^{j}\right) / w^{i}, \quad t_{y}^{ij}=\left(y^{i}-y^{j}\right) / h^{i}, \\ 
    &t_{w}^{ij}=\log \left(w^{i} / w^{j}\right), \quad 
    t_{h}^{ij}=\log \left(h^{i} / h^{j}\right), \\
    \nonumber
    &t_{x}^{ji}=\left(x^{j}-x^{i}\right) / w^{j}, \quad t_{y}^{ji}=\left(y^{j}-y^{i}\right) / h^{j}.
\end{align}

\begin{figure}[t]
    \centering
    \includegraphics[width=\linewidth]{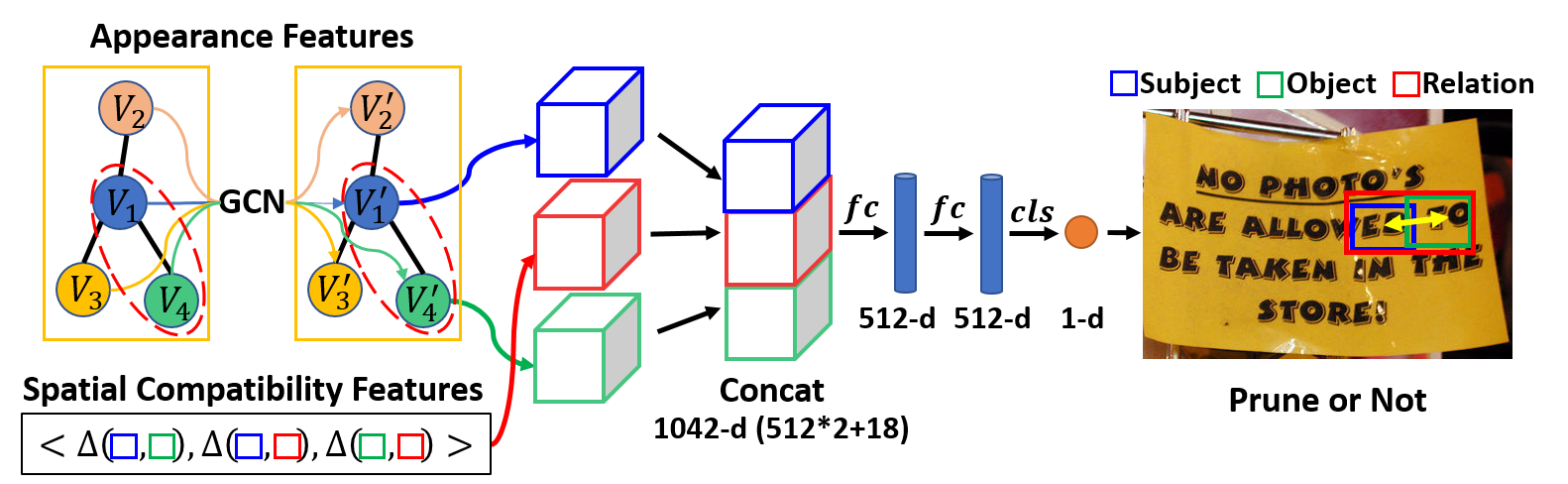}
    \caption{Architecture of the edge classifier for link relationship prediction.}
    \label{fig:RN}
\end{figure}

 As depicted in Fig.~\ref{fig:RN}, the edge classifier is a binary classifier which takes the feature representation of each edge as input and outputs whether the input edge should be pruned or not. It is implemented with a 2-hidden-layer MLP with 512 nodes at each hidden layer and 1 Sigmoid activation node at its output layer. As the text primitive graph is a directed graph, there could exist two edges with opposite directions linking a node pair. In our implementation, as long as there exists one edge linking a node pair after edge classification, we consider that there exists a link relationship between this node pair. 

To ease implementation, we directly use the ground-truth text primitives to construct the three text primitive subgraphs $\textbf{G}_{\textbf{P}_\textbf{2}}$, $\textbf{G}_{\textbf{P}_\textbf{3}}$, $\textbf{G}_{\textbf{P}_\textbf{4}}$ in the training stage. Moreover, we also add some synthesized ``non-text'' primitives into the subgraphs to improve the robustness of our GCN-based link prediction module to false alarms detected in the inference stage. Specifically, on each FPN pyramid level, we randomly sample a set of ``non-text'' pixels and assign each of them an artificial bounding box whose width and height are equal to the average width and height of all ground-truth text primitives respectively. Then, we assign all ``text'' and sampled ``non-text'' primitives an equal textness score (e.g., 1.0) and randomly shuffle them. After that, we perform a standard NMS algorithm with an IoU threshold of $0.3$ on them to generate a node set $\textbf{V}_{\textbf{P}_\textbf{k}} (\textrm{k} = 2, 3, 4)$, with which the corresponding subgraph is constructed by using the method proposed in Sec.~\ref{subsubsec:SGC}. In each subgraph, we label edges whose two nodes belong to a same text instance as positive, otherwise negative. During training, we ignore all the negative edges that connect two ``non-text'' nodes, and then adopt the OHEM \citep{shrivastava2016training} method to select an equal number of hard positive and hard negative samples to train our GCN-based link prediction module.

\begin{figure*}[t]
    \centering
    \includegraphics[width=\linewidth]{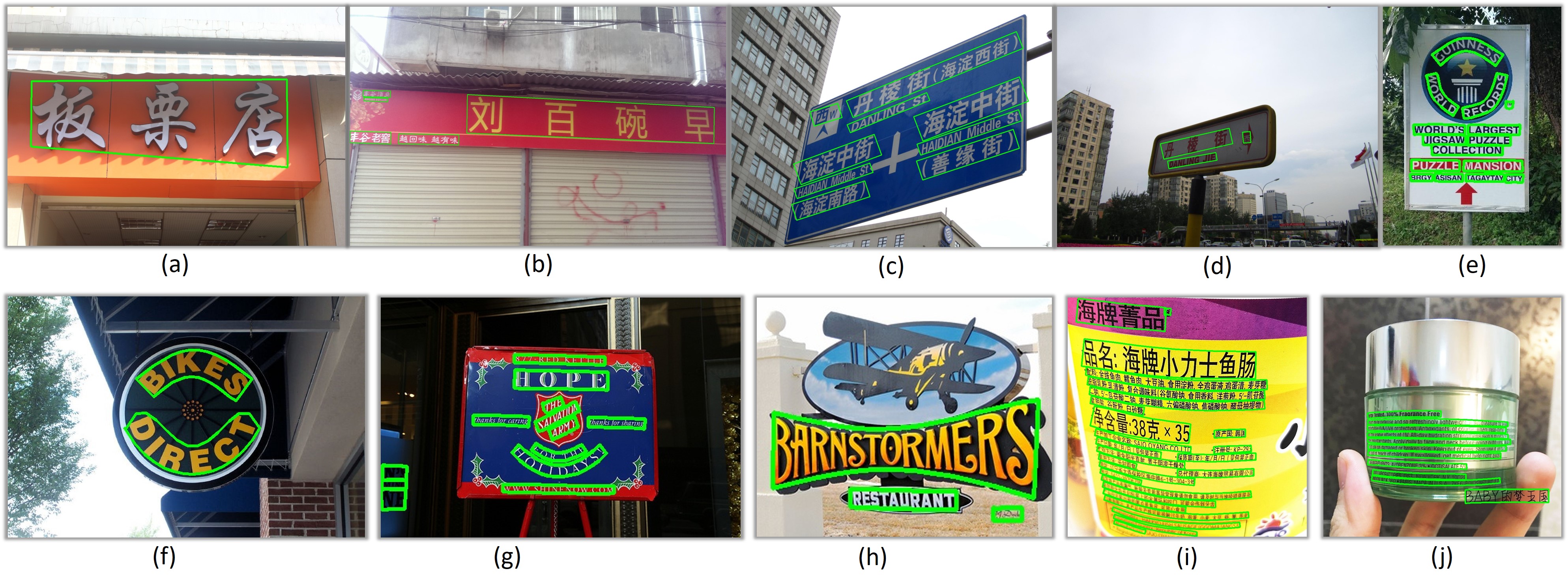}
    \caption{Qualitative detection results of ReLaText. (a-b) are from RCTW-17, (c-d) are from MSRA-TD500, (e-f) are from Total-Text, (g-h) are from CTW1500, and (i-j) are from DAST1500.}
    \label{fig:results}
\end{figure*}

\section{Loss Function}
\label{subsec:loss_func}
\textbf{Multi-task loss for text primitive detection.} There are two sibling output layers for each text primitive detection module, i.e., a textness score map prediction layer and a quadrilateral bounding box regression layer. The multi-task loss function is defined as follows:
\begin{equation}
    \mathcal{L}_{\textrm{TPD}_{\textrm{P}_i}} = \frac{1}{N}\sum_{j}\mathcal{L}_{cls}(c_j, c{_j}{^*}) + \frac{1}{N_{fg}}\sum_{k}\mathcal{L}_{reg}(t_k, t{_k}{^*}),
\end{equation}
where ${N}$ is the number of sampling pixels (including ${N_{fg}}$ positive ones), $c_j$ and $c{_j}{^*}$ are predicted and ground-truth labels for the $j$-th sampling pixel respectively, $\mathcal{L}_{cls}(c_j, c{_j}{^*})$ is a binary cross-entropy loss for classification tasks, $t_k$ and $t_k{^*}$ are predicted and ground-truth $8$-d normalized coordinate offsets as stated in \citep{zhong2019anchor} for the $k$-th positive sampling pixel, $\mathcal{L}_{reg}(t_k, t_k{^*})$ is a Smooth-$\textrm{L}_1$ loss \citep{ren2015faster} for regression tasks.

The total loss of the text primitive detection module is the sum of the losses of three scale-specific detection modules.

\textbf{Loss for edge classification.} Let $E$ denote the set of selected edges for edge classification training, $r_i$ and $r_i^*$ be the predicted and ground-truth labels for the $i$-th edge $e_i$, and $\mathcal{L}\left(r_i, r_i^*\right)$ be a binary cross-entropy loss for classification tasks. The loss function for edge classification is defined as follows:
\begin{equation}
    \mathcal{L}_{link} = \frac{1}{|E|}\sum_{e_i \in E}\mathcal{L}\left(r_i, r_i^*\right).
\end{equation}

\section{Experiments}
\label{sec:exps}

\subsection{Datasets and Evaluation Protocols}
\label{subsec:data}
To evaluate the performance of our proposed ReLaText, we conduct comprehensive experiments on five scene text detection benchmark datasets, including RCTW-17 \citep{shi2017icdar2017}, MSRA-TD500 \citep{yao2012detecting}, Total-Text \citep{ch2017total}, CTW1500 \citep{liu2019curved} and DAST1500 \citep{tang2019detecting}. We follow the evaluation protocols defined by the authors to make our results comparable to the ones from other methods. 

\textbf{RCTW-17} \citep{shi2017icdar2017} contains $8,034$ training images and $4,229$ testing images with scene texts in either Chinese or English. In this dataset, text instances are multi-oriented and labeled by quadrangles in text-line level.

\textbf{MSRA-TD500} \citep{yao2012detecting} contains $300$ training images and $200$ testing images, including English and Chinese scripts. Text instances in this dataset are multi-oriented and labeled by rotated rectangles in text-line level. Since the training images are too few, we follow the common practice of previous works to add $400$ more images from HUST-TR400 \citep{yao2014unified} to the training data. 

\textbf{Total-Text} \citep{ch2017total} contains $1,255$ training images and $300$ testing images, including $11,459$ text instances ($4,907$ are curved texts) in total. In this dataset, text instances are labeled in word-level with polygons.

\textbf{CTW1500} \citep{liu2019curved} contains $1,000$ training images and $500$ testing images. There are $10,751$ text instances in total, where $3,530$ are curved texts and at least one curved text per image. In this dataset, text instances are annotated in text-line level with 14-point polygons. 

\textbf{DAST1500} \citep{tang2019detecting} is a dense and arbitrary-shaped text detection dataset, which collects commodity images with detailed description of the commodities on small wrinkled packages from the Internet. It contains $1,038$ training images and $500$ testing images. Polygon annotations are given in text-line level.

\textbf{SynthText} \citep{gupta2016synthetic} is a synthetic dataset and consists of about 800k synthetic images generated by blending text instances rendered with random fonts, sizes, colors and orientations in natural images. As the training sets of MSRA-TD500, Total-Text, CTW1500 and DAST1500 are too small, we follow most previous methods \citep{long2018textsnake, tang2019detecting, baek2019character} to use this synthetic dataset with word-level annotations to pre-train our text detection models.

\begin{table*}[t]
    \setlength{\tabcolsep}{11.5pt}
    \centering
    \caption{Performance comparison on RCTW-17. ``L: 1500'' means that the longer side of each testing image is resized to be 1500 pixels. * indicates the results are from Liao et al. \cite{liao2018rotation}. ** lists the results of our model when testing with different image scales.}
    \label{tab:RCTW-17}
    \begin{tabular}{| c | c | c | c | c || c | c | c |}
        \hline
        \multirow{2}{*}{\textbf{Methods}} & \textbf{Image scale} & \multicolumn{3}{c||}{\textbf{RCTW-17}} & \multicolumn{3}{c|}{\textbf{**}} \\\cline{3-8}
         & \textbf{for testing} & \textbf{P(\%)} & \textbf{R(\%)} & \textbf{F(\%)} & \textbf{P(\%)} & \textbf{R(\%)} & \textbf{F(\%)} \\
        \hline\toprule
        \bottomrule\hline
        Official baseline \citep{shi2017icdar2017} & N/A & 76.0 & 40.4 & 52.8 & - & - & - \\\hline
        EAST$^{*}$ \citep{zhou2017east} & N/A & 59.7 & 47.8 & 53.1 & - & - & - \\\hline
        RRD \citep{liao2018rotation} & N/A & 72.4 & 45.3 & 55.7 & - & - & - \\\hline
        LOMO \citep{zhang2019look} & L:1024 & 80.4 & 50.8 & 62.3 & 77.3 & 57.8 & \textbf{66.1} \\\hline
        TextMountain \citep{zhu2018textmountain} & L:1500 & \textbf{80.8} & 55.2 & 65.6 & 75.9 & 61.7 & \textbf{68.1} \\\hline
        IncepText \citep{yang2018inceptext} & N/A & 78.5 & 56.9 & 66.0 & - & - & - \\
        \hline\toprule
        \bottomrule\hline
        \textbf{ReLaText (Proposed)} & L:1500 & 75.9 & \textbf{61.7} & \textbf{68.1} & & & \\
        \hline
    \end{tabular}
\end{table*}

\begin{table*}[t]
    \setlength{\tabcolsep}{9.25pt}
    \centering
    \caption{Performance comparison on MSRA-TD500. ``L: 900'' means that the longer side of each testing image is resized to be 900 pixels. ** lists the results of our model when testing with different image scales. Note that ``FPS'' is for reference only because the experimental environments are different.}
    \label{tab:MSRA-TD500}
    \begin{tabular}{| c | c | c | c | c | c || c | c | c |}
        \hline
        \multirow{2}{*}{\textbf{Methods}} & \textbf{Image scale} & \multicolumn{4}{c||}{\textbf{MSRA-TD500}} & \multicolumn{3}{c|}{\textbf{**}} \\\cline{3-9}
         & \textbf{for testing} & \textbf{P(\%)} & \textbf{R(\%)} & \textbf{F(\%)} & \textbf{FPS} & \textbf{P(\%)} & \textbf{R(\%)} & \textbf{F(\%)} \\
        \hline\toprule
        \bottomrule\hline
        EAST \citep{zhou2017east} & N/A & 87.3 & 67.4 & 76.1 & \textbf{13.2} & - & - & - \\\hline
        SegLink \citep{shi2017detecting} & $768 \times 768$ & 86.0 & 70.0 & 77.0 & 8.9 & 90.1 & 81.6 & \textbf{85.7} \\\hline
        Wu et al. \cite{wu2017self} & N/A & 77.0 & 78.0 & 77.0 & 4.0 & - & - & - \\\hline
        PixelLink \citep{deng2018pixellink} & $768 \times 768$ & 83.0 & 73.2 & 77.8 & 3.0 & 90.1 & 81.6 & \textbf{85.7} \\\hline
        TextSnake \citep{long2018textsnake} & $768 \times 1280$ & 83.2 & 73.9 & 78.3 & 1.1 & 86.9 & 83.0 & \textbf{84.9} \\\hline
        TextField \citep{xu2019textfield} & $768 \times 768$ & 87.4 & 75.9 & 81.3 & 5.2 & 90.1 & 81.6 & \textbf{85.7} \\\hline
        Lyu et al. \cite{lyu2018multi} & $768 \times 768$ & 87.6 & 76.2 & 81.5 & 5.7 & 90.1 & 81.6 & \textbf{85.7} \\\hline
        Tian et al. \cite{tian2019learning} & L:800 & 84.2 & 81.7 & 82.9 & 3.0 & 89.8 & 81.6 & \textbf{85.5} \\\hline
        CRAFT \citep{baek2019character} & L:1600 & 88.2 & 78.2 & 82.9 & 8.6 & 86.3 & 79.7 & \textbf{82.9}\\\hline
        SBD \citep{liu2019omnidirectional} & $1200 \times 1600$ & 89.6 & 80.5 & 84.8 & 3.2 & 86.3 & 79.7 & 82.9 \\
        \hline\toprule
        \bottomrule\hline
        \textbf{ReLaText (Proposed)} & L:900 & \textbf{90.5} & \textbf{83.2} & \textbf{86.7} & 8.3 &  &  &\\
        \hline
    \end{tabular}
\end{table*}

\begin{table*}[t]
    \setlength{\tabcolsep}{11.5pt}
    \centering
    \caption{Performance comparison on Total-Text. ``L: 1000'' means that the longer side of each testing image is resized to be 1000 pixels, while “S” is for shorter side. ** lists the results of our model when testing with different image scales.} 
    \label{tab:Total-Text}
    \begin{tabular}{| c | c | c | c | c || c | c | c |}
        \hline
        \multirow{2}{*}{\textbf{Methods}} & \textbf{Image scale} & \multicolumn{3}{c||}{\textbf{Total-Text}} & \multicolumn{3}{c|}{\textbf{**}} \\\cline{3-8}
         & \textbf{for testing} & \textbf{P(\%)} & \textbf{R(\%)} & \textbf{F(\%)} & \textbf{P(\%)} & \textbf{R(\%)} & \textbf{F(\%)} \\
        \hline\toprule
        \bottomrule\hline
        CENet \citep{liu2018detecting} & Original Size & 59.9 & 54.4 & 57.0 & 78.3 & 78.8 & \textbf{78.6} \\\hline
        Lyu et al. \cite{lyu2018mask} & S:1000  & 69.0 & 55.0 & 61.3 & 80.4 & 81.7 & \textbf{81.1} \\\hline
        TextSnake \citep{long2018textsnake} & $512 \times 512$  & 82.7 & 74.5 & 78.4 & 83.3 & 74.1 & \textbf{78.5} \\\hline
        MSR \citep{xue2019msr} & N/A & 83.8 & 74.8 & 79.0 & - & - & - \\\hline
        TextField \citep{xu2019textfield} & $768 \times 768$ & 81.2 & 79.9 & 80.6 & 86.4 & 81.8 & \textbf{84.0} \\\hline
        PSENet \citep{Wang2019psenet} & L:1280 & 84.0 & 78.0 & 80.9 & 81.8 & 82.3 & \textbf{82.0} \\\hline
        Mask R-CNN \citep{xie2019scene} & S:848 & 81.5 & 80.5 & 81.0 & 81.4 & 82.0 & \textbf{81.7} \\\hline
        SegLink++ \citep{tang2019detecting} & S:768 & 82.1 & 80.9 & 81.5 & 82.0 & 81.6 & \textbf{81.8} \\\hline
        LOMO \citep{zhang2019look} & N/A & \textbf{88.6} & 75.7 & 81.6 & - & - & -\\\hline
        CRAFT \citep{baek2019character} & L:1280 & 87.6 & 79.9 & 83.6 & 81.8 & 82.3 & 82.0 \\
        \hline\toprule
        \bottomrule\hline
        \textbf{ReLaText (Proposed)} & L:1000 & 84.8 & \textbf{83.1} & \textbf{84.0} &  &  & \\
        \hline
    \end{tabular}
\end{table*}

\begin{table*}[t]
    \setlength{\tabcolsep}{9.25pt}
    \centering
    \caption{Performance comparison on CTW1500. ``L: 800'' means that the longer side of each testing image is resized to be 800 pixels, while “S” is for shorter side. * indicates that the result is provided by author(s). ** lists the results of our model when testing with different image scales.} 
    \label{tab:CTW1500}
    \begin{tabular}{| c | c | c | c | c | c || c | c | c |}
        \hline
        \multirow{2}{*}{\textbf{Methods}} & \textbf{Image scale} & \multicolumn{4}{c||}{\textbf{CTW1500}} & \multicolumn{3}{c|}{\textbf{**}} \\\cline{3-9}
         & \textbf{for testing} & \textbf{P(\%)} & \textbf{R(\%)} & \textbf{F(\%)} & \textbf{FPS} & \textbf{P(\%)} & \textbf{R(\%)} & \textbf{F(\%)} \\
        \hline\toprule
        \bottomrule\hline
        CTD+TLOC \citep{liu2019curved} & $600 \times 1000$ & 77.4 & 69.8 & 73.4 & \textbf{13.3} & 86.3 & 83.3 & \textbf{84.8} \\\hline
        TextSnake \citep{long2018textsnake} & Original Size & 67.9 & 85.3 & 75.6 & - & 85.9 & 76.2 & \textbf{80.8} \\\hline
        LOMO \citep{zhang2019look} & L:512 & \textbf{89.2} & 69.6 & 78.4 & 4.4 & 87.6 & 78.2 & \textbf{82.6} \\\hline
        Tian et al. \cite{tian2019learning} & L:800 & 82.7 & 77.8 & 80.1 & - & 86.2 & 83.3 & \textbf{84.8} \\\hline
        Wang et al. \cite{wang2019arbitrary} & $720 \times 1280$ & 80.1 & 80.2 & 80.1 & 10.0 & 84.7 & 83.6 & \textbf{84.2} \\\hline
        SegLink++ \citep{tang2019detecting} & S:512 & 82.8 & 79.8 & 81.3 & - & 86.4 & 82.7 & \textbf{84.5} \\\hline
        TextField \citep{xu2019textfield} & $576 \times 576$ & 83.0 & 79.8 & 81.4 & - & 87.6 & 80.3 & \textbf{83.8} \\\hline
        MSR \citep{xue2019msr} & N/A & 85.0 & 78.3 & 81.5 & - & - & - & - \\\hline
        PSENet \citep{Wang2019psenet} & L:1280 & 84.8 & 79.7 & 82.2 & 3.9 & 84.1 & 83.4 & \textbf{83.8} \\\hline
        Mask R-CNN$^*$ \citep{huang2019mask} & S:512 & 83.8 & 81.7 & 82.7 & - & 86.4 & 82.7 & \textbf{84.5} \\\hline
        CRAFT \citep{baek2019character} & L:1024 & 86.0 & 81.1 & 83.5 & - & 84.8 & 82.8 & \textbf{83.8} \\
        \hline\toprule
        \bottomrule\hline
        \textbf{ReLaText (Proposed)} & L:800 & 86.2 & \textbf{83.3} & \textbf{84.8} & 10.6 & & & \\
        \hline
    \end{tabular}
\end{table*}

\begin{table}[t]
    \centering
    \caption{Performance comparison on DAST1500. * indicates the results are from \cite{tang2019detecting}.}
    \label{tab:DAST1500}
    \begin{tabular}{c | c c c}
        \hline
        Methods & P(\%) & R(\%) & F(\%)  \\
        \hline\hline
        SegLink$^{*}$ \citep{shi2017detecting} & 66.0 & 64.7 & 65.3 \\
        CTD+TLOC$^{*}$ \citep{liu2019curved} & 73.8 & 60.8 & 66.6 \\
        PixelLink$^{*}$ \citep{deng2018pixellink} & 74.5 & 75.0 & 74.7 \\
        SegLink++ \citep{tang2019detecting} & 79.6 & 79.2 & 79.4 \\
        \hline\hline
        \textbf{ReLaText (Proposed)} & \textbf{89.0} & \textbf{82.9} & \textbf{85.8} \\
        \hline
    \end{tabular}
\end{table}

\subsection{Implementation Details}
\label{subsec:details}
The weights of ResNet-50 related layers in the FPN backbone network are initialized with a pre-trained ResNet-50 model for the ImageNet classification task \citep{he2016deep}. The weights of newly added layers are initialized with a Gaussian distribution of mean $0$ and standard deviation $0.01$. Our ReLaText text detection models are trained in an end-to-end manner and optimized by a standard SGD algorithm with a momentum of $0.9$ and weight decay of $0.0005$. 

The number of training iterations and adjustment strategy of learning rate depend on the sizes of different datasets. Specifically, for RCTW-17, we use the provided $8,034$ training images for training and the model is trained for 400k iterations with a base learning rate of 0.004, which is divided by 10 at every 180k iterations. For the other four small scale datasets, we first use the SynthText dataset to pre-train a text detection model for 860k iterations with a base learning rate of 0.004, which is divided by 10 at every 387k iterations. The pre-trained model is then fine-tuned on the corresponding training sets of these four datasets, respectively. All these models are fine-tuned for 70k iterations with a base learning rate of 0.004, which is divided by 10 at every 32k iterations.

We implement ReLaText based on PyTorch\footnotemark\footnotetext{\url{https://pytorch.org/}}~v0.4.1 and conduct experiments on a workstation with 4 Nvidia V100 GPUs. In each training iteration, we sample one image for
each GPU. For each image, we sample a mini-batch of 128 text, 128 border and 128 background pixels for each text primitive detection module. And we select a mini-batch of 64 hard positive and 64 hard negative edges in each subgraph for edge classification training. We adopt a multi-scale training strategy during training. While keeping the aspect ratio, the shorter side of each selected training image is randomly rescaled to a number in $\{464, 592, 720, 848, 976\}$. Moreover, when training our text detector on MSRA-TD500 and RCTW-17, we also rotate training images by a random angle in $\{0^\circ, 90^\circ, 180^\circ, 270^\circ\}$ for data augmentation. 

In the testing phase, the detected text primitives are grouped into the whole text instances based on the predicted link relationships with a threshold of 0.7 by using the Union-Find algorithm. Then we use a polynomial curve fitting algorithm to fit the two long sides of a polygon (or a quadrilateral) that encloses all the grouped text primitives for each text instance. Final detection results from different pyramid levels are aggregated by a polygon NMS algorithm \citep{liu2019curved} with an IoU threshold of $0.2$. 

\subsection{Text Detection Performance Evaluation}
\label{subsec:TDPE}
In this section, we evaluate our ReLaText on RCTW-17, MSRA-TD500, Total-Text, CTW1500 and DAST1500. In all these experiments, a GCN with 3 layers is applied and all MLPs in the GCN have 2 layers. We compare the performance of our approach with other most competitive results on these five benchmark tasks. For fair comparisons, all results we reported are based on single-model and single-scale testing. 
\\
\textbf{Multilingual and multi-oriented text detection benchmarks.} We evaluate our approach on RCTW-17 and MSRA-TD500, which are both multilingual and multi-oriented text detection benchmarks. The longer sides of testing images are set as 1500 and 900 for RCTW-17 and MSRA-TD500, respectively. As shown in Table~\ref{tab:RCTW-17} and Table~\ref{tab:MSRA-TD500}, ReLaText outperforms the closest methods \citep{yang2018inceptext} and \citep{liu2019omnidirectional} substantially by improving F-score from 66.0\% to 68.1\% and 84.8\% to 86.7\% on RCTW-17 and MSRA-TD500, respectively. Moreover, as image scales used for inference affect significantly the text detection performance, we also compare our approach with other methods by using the same testing scales reported in their papers. Even without hyper-parameters tuning for those image scales, our approach can still achieve better results, which can further demonstrate the superior performance of our approach.
\\
\textbf{Curved text detection benchmarks.} We conduct experiments on two curved-text focused datasets, namely Total-Text and CTW1500, to evaluate the performance of ReLaText on detecting arbitrary-shaped texts in natural scene images. The longer sides of testing images are set as 1000 and 800 for Total-Text and CTW1500, respectively. Note that two latest methods \citep{xie2019scene, huang2019mask} adopt more powerful backbone networks, i.e., supervised pyramid context network in \citep{xie2019scene} and pyramid attention network in \citep{ huang2019mask} to push the text detection performance of Mask R-CNN on Total-Text and CTW1500, respectively. For fairer comparisons, we compare our approach with these two Mask R-CNN based methods by using the same ResNet50-FPN backbone network. As shown in Table~\ref{tab:Total-Text} and Table~\ref{tab:CTW1500}, ReLaText achieves the best F-score of 84.0\% on Total-Text and 84.8\% on CTW1500. Moreover, when using the same testing scales as other methods, ReLaText can still achieve better or comparable results. 
\\
\textbf{Dense and arbitrary-shaped text detection benchmark.} We further evaluate the performance of ReLaText on the newly proposed DAST1500 dataset, which contains a large number of dense and arbitrary-shaped text instances. Following other methods on this dataset, we resize the testing images to $768 \times 768$. As shown in Table~\ref{tab:DAST1500}, ReLaText achieves the best result of 89.0\%, 82.9\% and 85.8\% in precision, recall, and F-score, respectively, outperforming other competing approaches significantly. 
\\
\textbf{Inference time.} Based on our current implementation, ReLaText has an inference time of 0.29s, 0.12s, 0.31s, 0.09s and 0.23s per image when using a single V100 GPU for $L$=1500, $L$=900, $L$=1000, $L$=800 and $L$=768 on RCTW-17, MSRA-TD500, Total-Text, CTW1500 and DAST1500, respectively.
\\
\textbf{Qualitative results.} The superior performance achieved on the above five datasets demonstrates the effectiveness and robustness of ReLaText. As shown in Fig.~\ref{fig:results}, ReLaText can detect scene texts under various challenging conditions, such as non-uniform illumination, low contrast, low resolution, large inter-character spacings, small inter-line spacings and arbitrary shapes.

\begin{table}[t]
    \centering
    \caption{Ablation study of the depth of GCN.}
    \label{tab:gcn_depth}
    \begin{tabular}{c | c c c}
        \hline
        Depth of GCN & P(\%) & R(\%) & F(\%)  \\
        \hline\hline
        1 & 85.5 & 82.5 & 84.0 \\
        2 & 86.2 & 82.4 & 84.2 \\
        3 & 86.2 & \textbf{83.3} & \textbf{84.8} \\
        4 & \textbf{86.5} & 82.5 & 84.5 \\
        \hline
    \end{tabular}
\end{table}

\begin{table}[t]
    \centering
    \caption{Ablation study of the depth of each MLP in GCN.}
    \label{tab:mlp_depth}
    \begin{tabular}{c | c c c}
        \hline
        Depth of each MLP & P(\%) & R(\%) & F(\%)  \\
        \hline\hline
        1 & 85.0 & 82.8 & 83.9 \\
        2 & \textbf{86.2} & \textbf{83.3} & \textbf{84.8} \\
        3 & 86.0 & 82.9 & 84.4 \\
        4 & 86.0 & 82.5 & 84.2 \\
        \hline
    \end{tabular}
\end{table}

\begin{table*}[t]
    \setlength{\tabcolsep}{11pt}
    \centering
    \caption{Comparison of different text-line grouping strategies. ``Connectivity'', ``Linkage'', ``RN based link relationship'', ``GCN based link relationship'' denote grouping based on 8-neighborhood pixel connectivity, 8-neighborhood pixel linkage, link relationships predicted by Relation Network and GCN, respectively.}
    \label{tab:connectivity-linkage-relation}
    \begin{tabular}{| c | c | c | c || c | c | c |}
        \hline
        \multirow{2}{*}{\textbf{Methods}} & \multicolumn{3}{c||}{\textbf{CTW1500}} & \multicolumn{3}{c|}{\textbf{RCTW-17}} \\\cline{2-7}
         & \textbf{P(\%)} & \textbf{R(\%)} & \textbf{F(\%)} & \textbf{P(\%)} & \textbf{R(\%)} & \textbf{F(\%)} \\
        \hline\toprule
        \bottomrule\hline
        Connectivity (e.g., \cite{wu2017self, long2018textsnake, xue2019msr}) & 83.6 & 81.6 & 82.6 & 76.0 & 58.6 & 66.2 \\\hline
        Linkage (e.g., \cite{shi2017detecting, deng2018pixellink, tang2019detecting}) & 84.0 & 81.4 & 82.7 & \textbf{76.8} & 58.1 & 66.2 \\\hline
        RN based link relationship \citep{ma2019relation} & 85.0 & 82.5 & 83.8 & 76.4 & 59.6 & 67.0 \\\hline
        \textbf{GCN based link relationship (ReLaText)} & \textbf{86.2} & \textbf{83.3} & \textbf{84.8} & 75.9 & \textbf{61.7} & \textbf{68.1} \\
        \hline
    \end{tabular}
\end{table*}

\begin{figure*}[t]
    \centering
    \includegraphics[width=\linewidth]{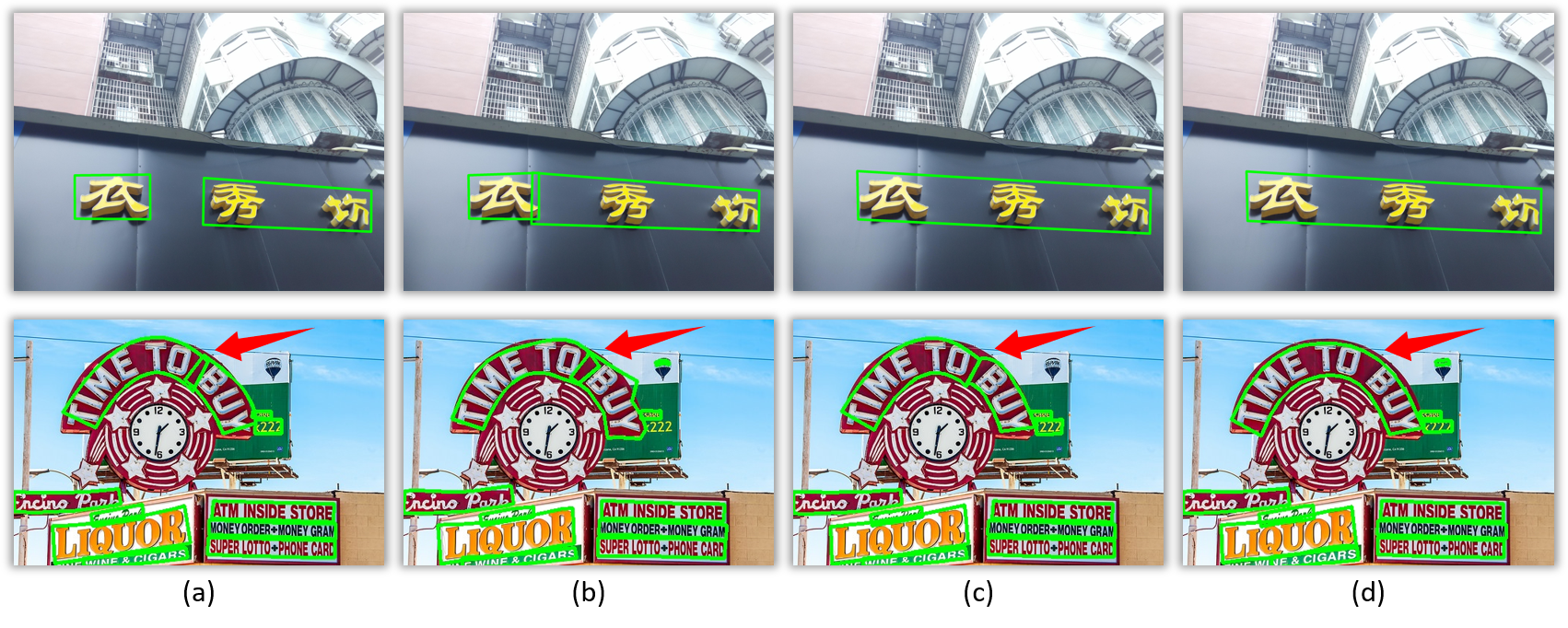}
    \caption{Some comparison examples from different text-line grouping approaches: (a) Connectivity; (b) Linkage; (c) RN based link relationship prediction; (d) GCN based link relationship prediction (ReLaText).}
    \label{fig:comparison_examples}
\end{figure*}

\subsection{Ablation Stuy}
\label{subsec:discussion}
\textbf{Influence of the network depth.} We investigate the influences of the depths of GCN and MLPs in GCN layers to text detection accuracies, respectively. The experimental results are shown in Table~\ref{tab:gcn_depth} and Table~\ref{tab:mlp_depth}. Since less parameters harm the performance and more parameters do not bring extra gains, setting the depth of GCN as 3 and the depth of each MLP as 2 exhibits a good trade-off between performance and efficiency for our approach.

\textbf{Effectiveness of visual relationship prediction based text-line grouping.} We compare our two visual relationship prediction based text-line grouping methods with two previously widely used text-line grouping methods, which are based on 8-neighborhood pixel connectivity and linkage respectively, on CTW1500 and RCTW-17 to demonstrate the effectiveness of the visual relationship prediction based problem formulation for text detection. In our conference paper \citep{ma2019relation}, we used a relation network to do link relationship prediction. In this work, we propose to exploit a GCN model to further improve the accuracy of link relationship prediction. For a fair comparison, we replace our line grouping module with the 8-neighborhood pixel connectivity and linkage based methods in our codes and tune the hyper-parameters carefully to get their best possible results. The quantitative results are given in Table~\ref{tab:connectivity-linkage-relation}, from which we can observe that the performance of pixel connectivity and linkage based methods are obviously inferior to our link relationship based methods, i.e., 82.6\% and 82.7\% vs. 83.8\% and 84.8\% on CTW1500, 66.2\% and 66.2\% vs. 67.0\% and 68.1\% on RCTW-17, which demonstrates the superiority of visual relationship prediction based text-line grouping.

\textbf{Effectiveness of GCN-based link relationship prediction.} We compare the GCN-based link relationship prediction with our previous relation network based link relationship prediction to demonstrate the effectiveness of GCN for link relationship prediction. As shown in the last two rows in Table~\ref{tab:connectivity-linkage-relation}, exploiting GCN to predict link relationship is more effective than relation network, and the F-score is improved substantially from 83.8\% to 84.8\% on CTW1500 and from 67.0\% to 68.1\% on RCTW-17 respectively. Some comparison examples are shown in Fig.~\ref{fig:comparison_examples}. Based on our observations, ReLaText can leverage context information more effectively to improve link relationship prediction accuracy, leading to improved robustness to text instances with large inter-character or small inter-line spacings.

\begin{figure}[t]
    \centering
    \includegraphics[width=0.9\linewidth]{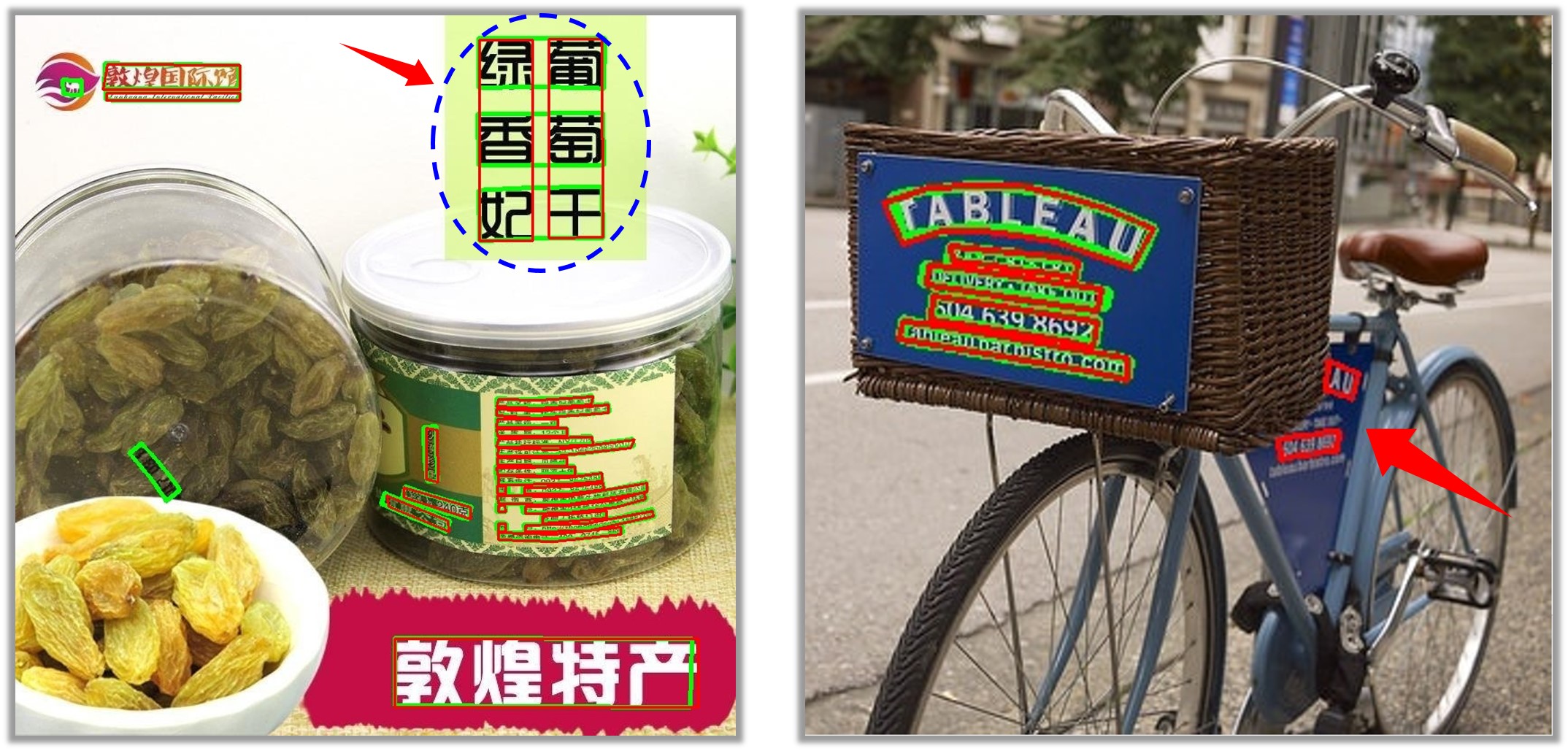}
    \caption{Some failure examples. Ground-truths are depicted in red, and detection results are depicted in green.}
    \label{fig:failure_cases}
\end{figure}

\subsection{Limitations of Our Approach}
\label{subsec:limitations}
Although our ReLaText shows superior capability in most scenarios as demonstrated in the previous experiments, it still fails in some difficult cases such as text-lines with ambiguous layouts and extremely small texts. Some failure examples are presented in Fig.~\ref{fig:failure_cases}. Note that these difficulties are common challenges for other state-of-the-art methods. Finding effective solutions to these problems will be our future work.

\section{Conclusion and Future Work}
\label{sec:conclusion}
In this paper, we introduce a new arbitrary-shaped text detection approach by using a GCN-based visual relationship detection framework to solve the challenging text-line grouping problem. As GCN can effectively leverage context information to improve link prediction accuracy, our visual relationship prediction based text-line grouping approach can achieve better text detection accuracy than previous local pixel connectivity based methods, especially when dealing with text instances with large inter-character or very small inter-line spacings. Consequently, the proposed ReLaText has achieved state-of-the-art performance on five public benchmark datasets, namely RCTW-17, MSRA-TD500, Total-Text, CTW1500 and DAST1500. 

For future work, we will explore other kinds of relationships, such as a “duplicate” relationship to suppress duplicate text instances, to improve text detection accuracy. We will also study how to integrate our text detector into other visual relationship detection based document understanding systems to help improve their accuracies. Furthermore, we will develop better text primitive detection methods to improve the robustness of our text detector to extremely small or extremely large text instances.

\bibliography{ReLaText.bbl}
\end{document}